\documentclass[letterpaper, 10 pt, journal, twoside]{IEEEtran}

\usepackage{cite}

\usepackage[pdftex]{graphicx}
\usepackage{multicol}
\usepackage{placeins}
\usepackage{float}

\usepackage{amsmath}
\usepackage{amssymb}
\usepackage{mathtools}
\usepackage{dsfont}
\usepackage{bbm}
\usepackage{bm}
\usepackage{xcolor}   
\usepackage{siunitx}
\usepackage{booktabs}
\usepackage{multirow}
\usepackage{caption}
\newcommand{\vect}[1]{\bm{#1}}

\newcommand{\figref}[1]{Fig.~\ref{#1}}
\newcommand{\secref}[1]{Sec.~\ref{#1}}

\newcommand{\tabref}[1]{Tab.~\ref{#1}}

\definecolor{blue}{rgb}{0.1216, 0.4667, 0.7059}
\definecolor{magenta}{rgb}{0.502, 0.0, 0.502}
\definecolor{green}{rgb}{0.2225, 0.5275, 0.2225}
\definecolor{orange}{rgb}{1.0, 0.498, 0.055}
\definecolor{red}{RGB}{184,54,30}
\definecolor{rebuttal}{RGB}{0,0,0}
\definecolor{cyan}{RGB}{57,227,194}

\begin{document}

\title{Olaf: Bringing an Animated Character to Life in the Physical World}



\author{David Müller$^{*}$, Espen Knoop$^{*}$, Dario Mylonopoulos, Agon Serifi,\\Michael A. Hopkins, Ruben Grandia, and Moritz Bächer%
\thanks{$^{*}$Joint first authors.}%
\thanks{The authors are with Disney Research Imagineering.}%
}

\maketitle


\begin{abstract}

Animated characters often move in non-physical ways and have proportions that are far from a typical walking robot. This provides an ideal platform for innovation in both mechanical design and stylized motion control. In this paper, we bring Olaf to life in the physical world, relying on reinforcement learning guided by animation references for control.
To create the illusion of Olaf's feet moving along his body, we hide two asymmetric legs under a soft foam skirt. 
To fit actuators inside the character, we use spherical and planar linkages in the arms, mouth, and eyes.
Because the walk cycle results in harsh contact sounds, we introduce additional rewards that noticeably reduce impact noise.  
The large head, driven by small actuators in the character's slim neck, creates a risk of overheating, amplified by the costume. To keep actuators from overheating, we feed temperature values as additional inputs to policies, introducing new rewards to keep them within bounds. 
We validate the efficacy of our modeling in simulation and through experiments on the robot.
\end{abstract}


\section{Introduction}

\IEEEPARstart{T}{he} field of legged robotics has traditionally been driven by goals centered on functionality, robustness, and efficiency~\cite{hurst2019walk}. 
This focus has enabled impressive dynamic capabilities, from hiking up mountains~\cite{miki2022learning} to traversing challenging terrain~\cite{rudin2025parkour, zhu2025artemis}. However, as robots enter domains that involve direct human participation, such as entertainment~\cite{fujita2001aibo, liu2025screen, grandia2024bdx, kozima2009keepon} and companionship~\cite{robinson2013psychosocial}, functional performance alone is no longer sufficient. In these settings, believability and character fidelity become central, shifting the engineering objective and imposing strict aesthetic constraints. 

\begin{figure}[t]
    \centering
        \includegraphics[width=\linewidth]{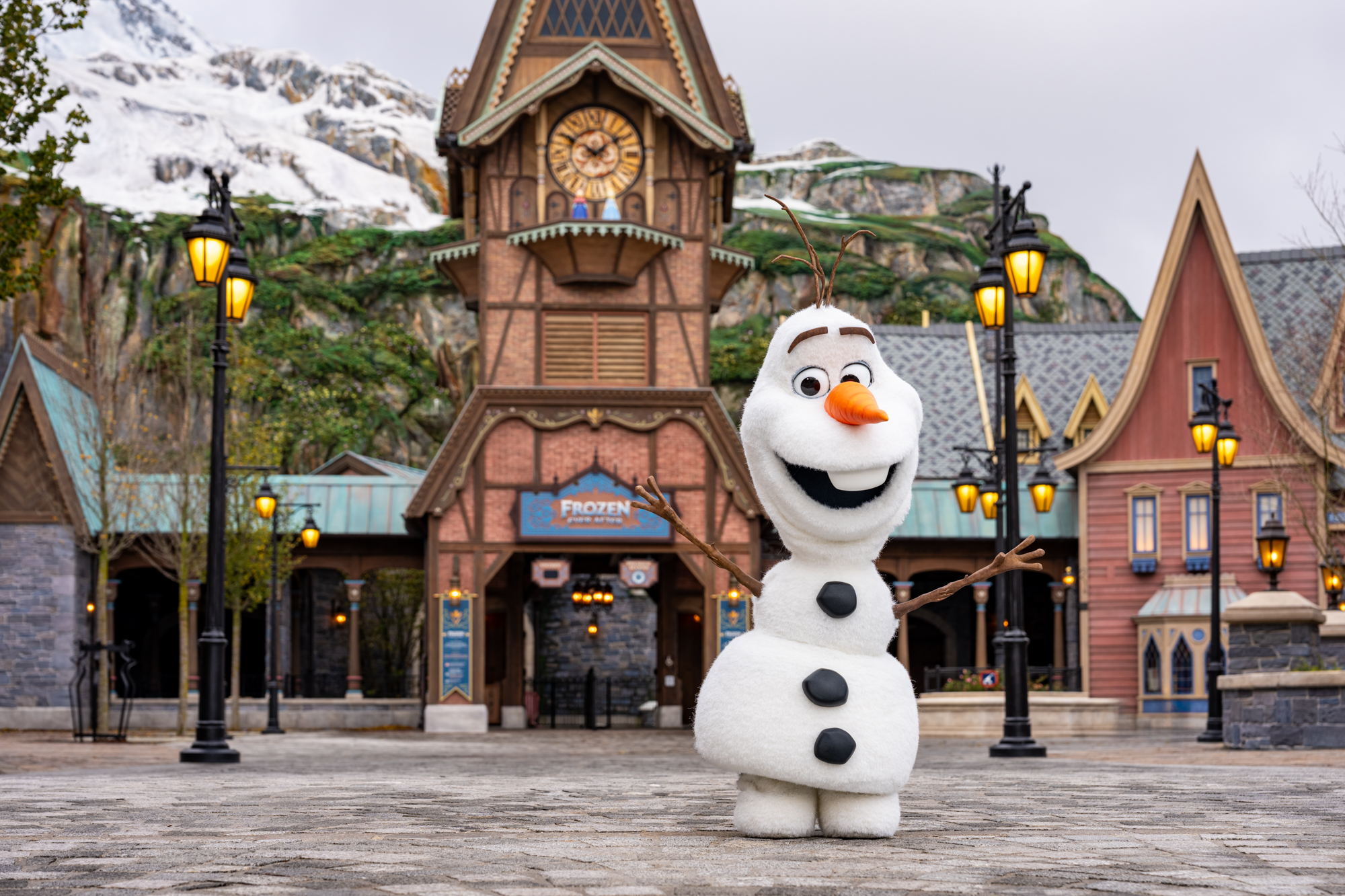}
    \caption{\textbf{Olaf Robot.}}
    \label{fig:teaser}
\end{figure}

In this paper, we focus on bringing Olaf, an animated character, into the physical world (see~\figref{fig:teaser}). Unlike most robots, Olaf has a large, heavy head, small snowball feet, as well as a non-physical motion style. These characteristics impose strict constraints on both mechanical design and motion control, including the need to hide all mechanical components within tight spatial constraints. At the same time, even small inconsistencies, such as rough foot impacts or jitter, can break the character’s lifelike appearance. 
This sensitivity makes it particularly challenging to develop a convincing robotic representation of an animated character.

We address the mechanical challenges with a compact design that is fully hidden beneath a costume. To achieve the required mobility of the leg mechanism within the limited volume, we adopt an asymmetric leg design and conceal it under a soft foam skirt. This skirt also shapes the costume and creates the illusion of Olaf’s feet moving freely beneath his body. To preserve Olaf’s visual appearance, we rely on small actuators and remote actuation through spherical and planar linkages. These are used throughout the character in the arms, mouth, and eyes. 

The control of Olaf is based on Reinforcement Learning (RL), with imitation rewards centered around animation references. The combination of a large head and small actuators placed inside the slim costumed neck introduces a risk of overheating. To address this, we propose a policy that takes actuator temperatures as inputs. We incorporate a thermal actuator model into our simulation and formulate a reward that encourages thermal safety. Because harsh robotic footsteps break the character’s believability, we introduce a reward that noticeably reduces impact noise. Our reward formulation also includes joint limit constraints and a penalty for collisions between the feet. We complement the RL policies with classical control and system identification to actuate the mechanical linkages in the arms, eyes, and mouth. 

Succinctly, our main contributions are:
\begin{itemize}
    \item \textbf{Mechatronic design:} a compact, scale-accurate design of Olaf, featuring a novel asymmetric six-degrees-of-freedom (6-DoF) leg mechanism and the integration of remotely actuated spherical, planar, and spatial linkages to achieve high-fidelity, expressive motion in the arms, mouth, and eyes.
    \item \textbf{Thermal-aware policy:} a control policy that incorporates actuator temperature as an input and learns to prevent overheating through our proposed reward formulation.
    \item \textbf{Impact reduction reward:} a reward that substantially reduces footstep noise, helping to preserve the character’s believability.
\end{itemize}


\section{Related Work}

Most legged robots take inspiration from Nature through their anthropomorphic~\cite{sakagami2002intelligent, oh2006design, gouaillier2009mechatronic, liao2025berkeley, lohmeier2009humanoid} or zoomorphic~\cite{seok2014design, hutter2016anymal, reher2019dynamic} morphologies, and their design and control are predominantly guided by functional requirements~\cite{saloutos2023design, seok2014design}. Recent advancements have pushed these boundaries through transformative mechanical designs, such as avian-inspired clutching linkages for energy efficiency~\cite{badri2022birdbot} or multimodal skeletal structures that mimic small-scale mammals~\cite{shi2022development}.
In this work, we instead focus on creating a robotic character, based on an artistic reference --- bringing a well-known animated character into the physical world. The robot should freely walk and move like the animated character, with functional aspects such as energy efficiency and power density becoming secondary considerations.

Related work~\cite{grandia2024bdx} created a \emph{new} robotic character and presented a pipeline for animating and controlling it. Our work differs in that we are bringing an \emph{existing} animated character to life, which requires navigating tradeoffs of functionality and believability within a tight design envelope. While the Cosmo robot~\cite{liu2025screen} represents an existing character from a movie, we focus on a non-robotic, costumed character with less favorable proportions.

Bipedal and quadrupedal robots often have actuators at the joints~\cite{sakagami2002intelligent, hutter2016anymal}, or implement remote actuation through linkages~\cite{hutter2012starleth, reher2019dynamic, wensing2017proprioceptive}. Because the two legs of Olaf are hidden in its small main body, we depart from a traditional symmetric design, to maximize the workspace. We also use linkages to place actuators where there is space~\cite{maloisel2025versatile,maloisel2023optimal}, including spherical linkages at the shoulders.

For control, we rely on policies trained using RL that are conditioned on high-level control inputs. During runtime, these high-level control inputs are computed by a real-time character animation engine~\cite{grandia2024bdx}. RL has led to tremendous progress in robust locomotion~\cite{peng2017deeploco, hwangbo2019learning, li2021reinforcement, xie2018feedback}, imitation learning~\cite{peng2018deepmimic, peng2020learning, serifi2024vmp, he2025asap}, and navigation of complex environments~\cite{hoeller2023anymal, he2025attention}. More recently, RL has increasingly been used to account for additional real-world effects, ranging from energy losses in actuators~\cite{bjelonic2025towards} to impact-minimizing strategies that allow quieter locomotion~\cite{zhang2025quietpaw, watanabe2025learning}. We build on these ideas to reduce Olaf’s stepping sound while preserving its characteristic gait, and further introduce thermal-aware policies with applications beyond our specific use case.

\section{Workflow}
\label{sec:workflow}

\begin{figure}[t]
    \centering
        \includegraphics[width=\linewidth]{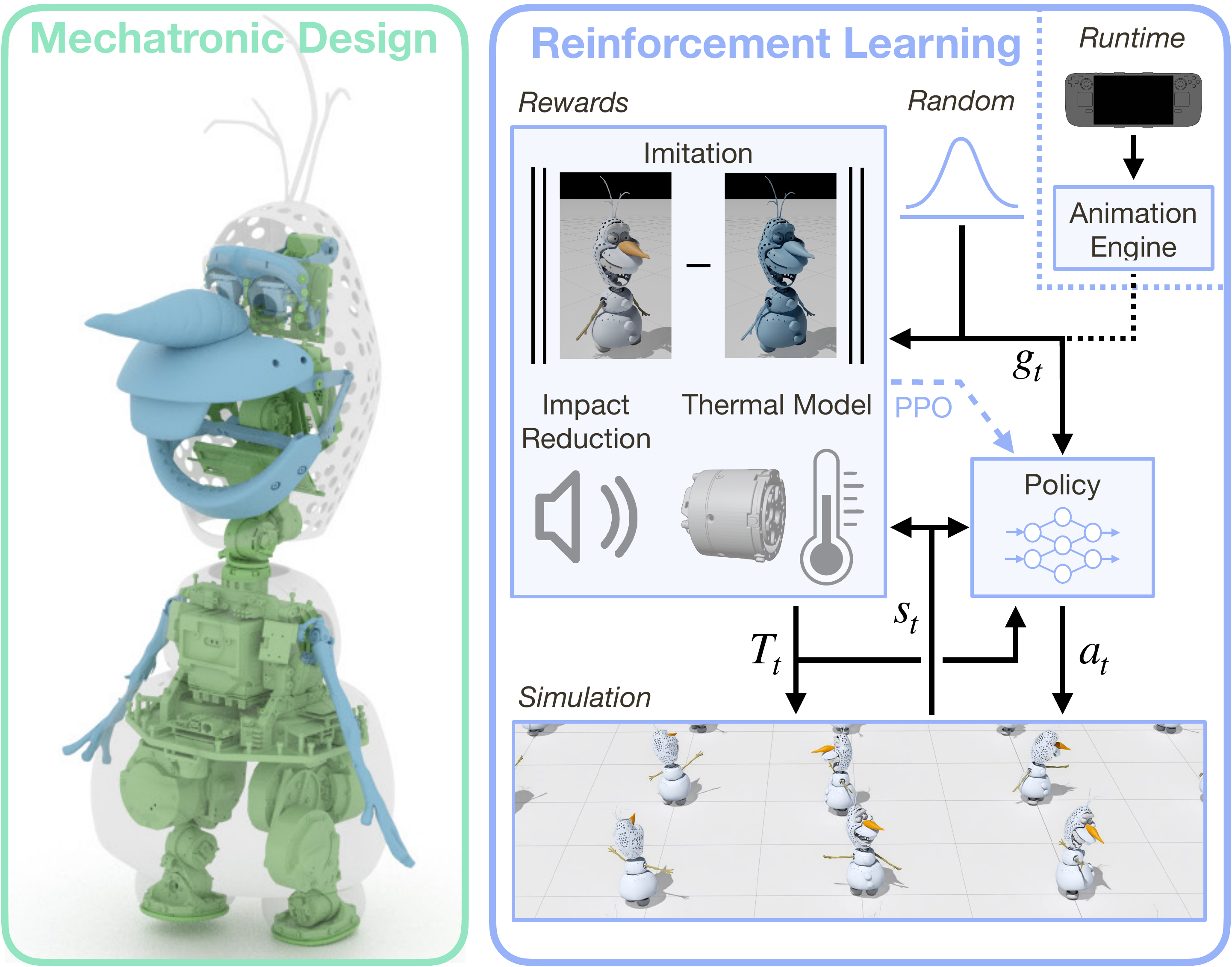}
    \caption{\textbf{Mechatronic Design and RL-based Control.} We separate the \textcolor{green}{articulated backbone} from the \textcolor{blue}{show functions}. The backbone is controlled via policies conditioned on the high-level control input $\vect{g}_t$ and trained using a combination of imitation, overheating, and impact rewards. During training, the control inputs are randomized, whereas at runtime, the Animation Engine generates control inputs from puppeteering commands.}
    \label{fig:overview}
\end{figure}

Our goal is to realize a robotic representation of Olaf, an animated character with characteristic, non-physical movements, and proportions atypical for robots.
To meet these strict functional and creative requirements, the initial mechanical design began with the main backbone with actuated legs and neck (shown in green in Fig.~\ref{fig:overview}). For animation authoring, we maintained an animation rig and animation references with the same degrees of freedom. To rapidly explore where to best place actuated degrees of freedom within the envelope of the character, we iteratively trained policies for standing and walking, evaluating the expressiveness of the character skeleton in simulation. In a second phase, we added mechanical show functions---elements that drive expressive behavior such as the arms, mouth, eyes, and eyebrows (in blue) without affecting system dynamics. See \secref{sec:mechatronic_design} for a description of the mechatronic design.

Control is divided into two layers: the articulation backbone, controlled by RL policies (\secref{sec:reinforcement_learning}), and the show functions, using classical control methods (\secref{sec:show_functions}). The RL simulation model is derived from the mechanical design and augmented with actuator temperature dynamics. Policies are trained using a reward function that augments imitation terms for accurate tracking of kinematic references with penalties enforcing physical limits, such as joint ranges and actuator temperatures. We train separate walking and standing policies, each conditioned on control inputs $\vect{g}_t$ for animation tracking and interactive control.

At runtime (\secref{sec:runtime}), Olaf is puppeteered via a remote interface. Following the design in~\cite{grandia2024bdx}, the \emph{Animation Engine} processes these commands to switch policies, trigger animations and audio, and provide interactive joystick control.

\section{Mechatronic Design}
\label{sec:mechatronic_design}

\begin{figure}
    \centering
    \includegraphics[width=\linewidth]{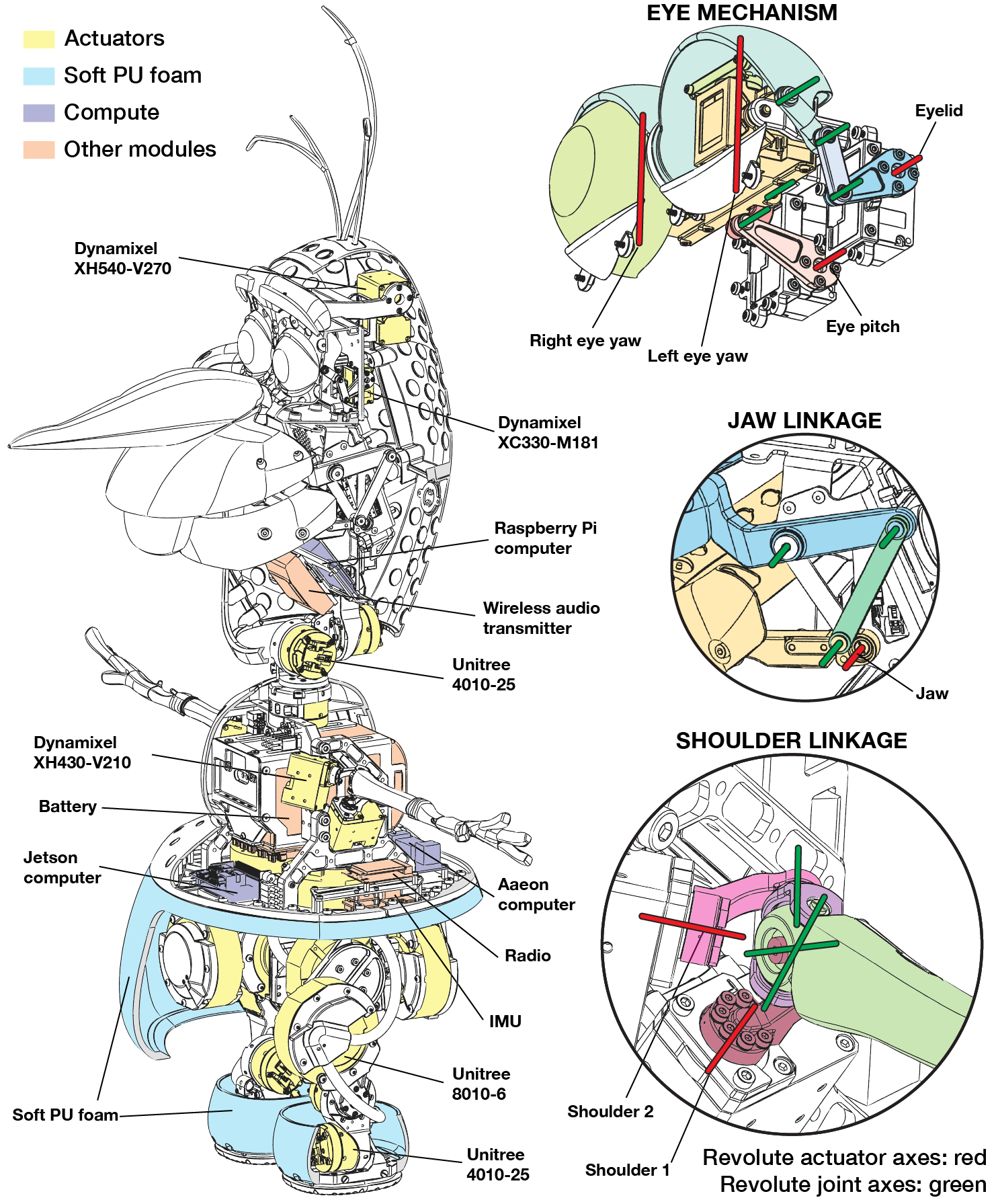}
    \caption{\textbf{Mechatronic Design.} Shells and skirt have been cut away to show the interior. Note that the costume is not shown. }
    \label{fig:section_view}
\end{figure}

The presented robotic representation of Olaf stands \SI{88.7}{\centi\meter} tall without hair and weighs \SI{14.9}{\kilogram}. It has $25$ degrees of freedom in total; $6$ per leg, $2$ per shoulder, $3$ in the neck, $1$ in the jaw, $1$ in the eyebrow, and $4$ in the mechanical eyes. See \figref{fig:section_view} for an annotated cutaway view of the robot. We use Unitree and Dynamixel actuators as indicated in the figure, and on-board compute is provided by 3 computers.
The form factor of the character presents several design challenges, discussed next. 

\paragraph*{Compact Design Envelope}

For the animated character, the two feet are free-floating snowballs beneath the body without visible legs. To emulate this, the robot design largely conceals the legs within the lower body, constraining the motion envelope of each leg to be within this bounding volume. To this end, we implement a novel 6-DoF leg design where one leg is inverted, such that the left leg has a rear-facing hip roll actuator and a forward knee, and the right leg has a forward hip roll actuator and a rear-facing knee (see~\figref{fig:section_view}). This configuration mitigates collisions between the two hip roll actuators and between the knees as the legs rotate in yaw. Furthermore, as the two legs are identical, instead of being mirrored, the part count is reduced.

The robot requires a 2-DoF shoulder; however, the limited space prevents placing the actuators at the joint. Instead, we place actuators within the torso and drive the shoulder through a spherical 5-bar linkage. See \figref{fig:section_view}, and also the supplemental video. For the mouth, a single actuator drives both the upper and lower jaw. The lower jaw is actuated directly, and the upper jaw is coupled to this through a 4-bar linkage. The mechanical eyes have independent direct-drive eye yaw, along with eye pitch and eyelid, which are both remotely actuated through 4-bar linkages. All other joints are direct-drive.

\paragraph*{Soft Shells} To conceal the legs within the lower snowball, while not overly restricting the range of motion, the lower snowball is designed as a flexible ``skirt'' made from polyurethane (PU) foam. This offers sufficient structure to maintain its shape while allowing deflection to accommodate larger leg movement, such as during recovery steps. The foot snowballs are constructed similarly. The flexible foam also absorbs impacts, reducing the severity of falls.

\paragraph*{Costuming and Appendages}

The costume is made of 4-way stretch fabric that can stretch both horizontally and vertically, allowing it to conform to the robot and its movements. A semi-rigid ``boning'' structure ensures that the costume maintains its shape over the cavity below the mouth. Around the eyes and the mouth, snap fasteners and magnets hold the costume in place. The arms, nose, buttons, eyebrows, and hair are held in place with magnets. This enables them to affix atop the costume (nose, buttons), and to break away in case of a fall or impact to mitigate damage.

\section{Reinforcement Learning}
\label{sec:reinforcement_learning}

Building on previous work in character control~\cite{grandia2024bdx}, we use a separate walking and standing policy, each formulated as an independent RL problem tailored to its specific motion regime. At each time step, the agent produces an action $\vect{a}_t$ according to a policy $\pi(\vect{a}_t | \vect{s}_t, \vect{g}_t)$, conditioned on the observable state $\vect{s}_t$ and control input $\vect{g}_t$. The environment then returns the next state $\vect{s}_{t+1}$ and a scalar reward $r_t = r(\vect{s}_t, \vect{a}_t, \vect{s}_{t+1}, \vect{g}_t)$, which encourages accurate imitation of artist-defined kinematic motions while maintaining dynamic and robust balance.

To achieve invariance to the robot's global pose and enable smooth transitions between policies, we use the path frame concept introduced in~\cite{grandia2024bdx}. The path frame state at time $t$ is
\begin{equation}
\vect{p}^{\mathrm{PF}}_t \coloneq \big( x^{\mathrm{PF}}_t,\, y^{\mathrm{PF}}_t,\, \psi^{\mathrm{PF}}_t \big),
\end{equation}
where $(x^{\mathrm{PF}}_t, y^{\mathrm{PF}}_t)$ denote the horizontal position and $\psi^{\mathrm{PF}}_t$ the yaw orientation.  

\begin{figure}[t]
    \centering
    \includegraphics[width=0.65\linewidth]{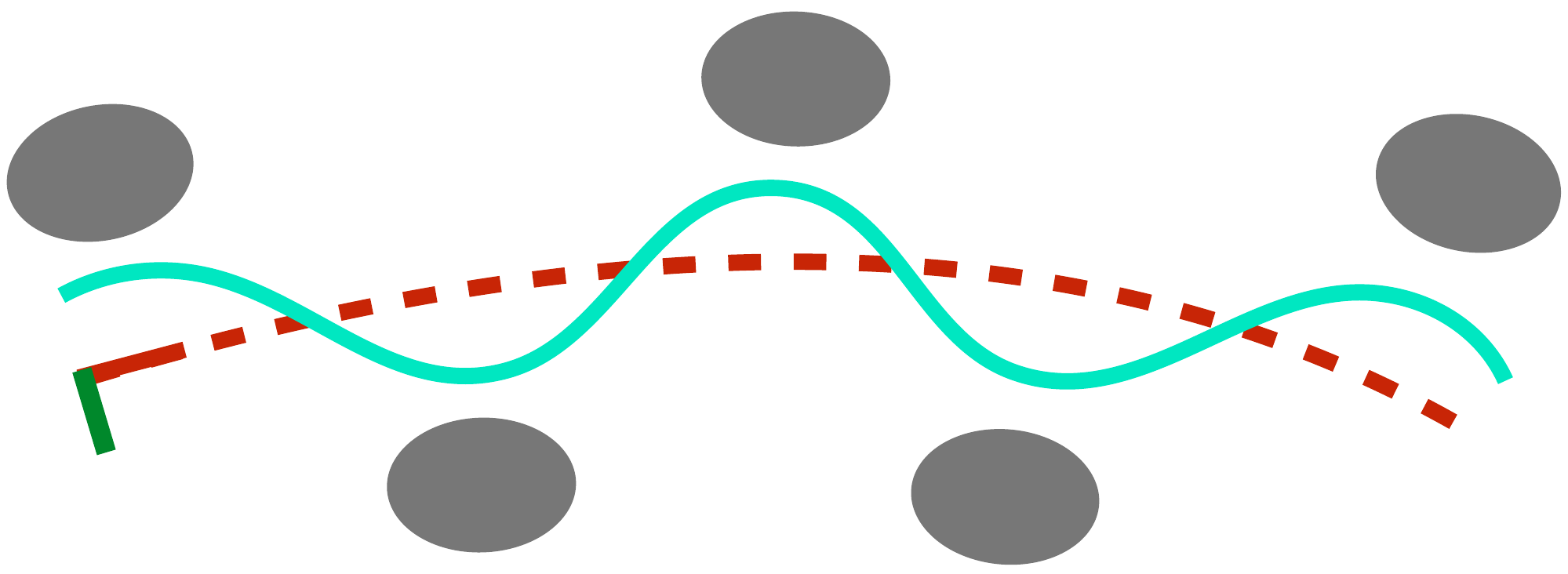}
    \caption{\textbf{Path Frame.} Visualization of the \textcolor{red}{path-frame} concept from~\cite{grandia2024bdx} and the robot's \textcolor{cyan}{center of mass}.}
    \label{fig:path_frame}
\end{figure}

During walking, the path frame advances by integrating the commanded path velocity $\vect{v}_t^{\text{PF}}$ (see Fig.~\ref{fig:path_frame}). During standing, the frame slowly converges toward the midpoint between the feet. Quantities expressed relative to this frame are denoted with the superscript $\mathcal{P}$. To prevent excessive deviation from the robot's motion, the path frame is constrained to remain within a bounded distance of the torso.

\subsection{Animation Reference}
\label{sec:animation_reference}

We start with animation references for walking and standing, created by artists using animation tools. 
These are conditioned on a control input $\vect{g}_t$, defined in \eqref{eq:motion_generation}. Notably, we use a gait generation tool \cite{motion_engine} to design stylized walk cycles with heel-toe motion, capturing Olaf’s characteristic gait. We demonstrate the importance of this feature in~\secref{sec:tracking}. Based on these references, the full kinematic target state $\vect{x}_t$ is obtained through a generator function $f(\cdot)$ that maps the path-frame state $\vect{p}^{\mathrm{PF}}_t$ and the policy-dependent control input $\vect{g}_t$ to the kinematic target using interpolation and path-frame alignment. This mapping is expressed as

\begin{equation}
    \begin{aligned}
    \vect{x}_t &\coloneq (\vect{p}^\mathcal{P}_t, \vect{\theta}^\mathcal{P}_t, \vect{v}^\mathcal{P}_t, \vect{\omega}^\mathcal{P}_t,\vect{q}_t, \dot{\vect{q}}_t, c^L_t, c^R_t), \\
    \vect{g}_t &\coloneq
    \begin{cases}
        (\hat{\vect{q}}^{\text{neck}}_t, \hat{\vect{\theta}}_t, \hat{\vect{p}}_{z,t}) & \text{standing},\\
        (\hat{\vect{q}}^{\text{neck}}_t, \hat{\vect{v}}_t^{\text{PF}}) & \text{walking},
    \end{cases} \\
    \vect{x}_t &=
    \begin{cases}
        f(\vect{p}^{\mathrm{PF}}_t, \vect{g}_t) & \text{standing},\\
        f(\vect{p}^{\mathrm{PF}}_t, \vect{g}_t, \phi_t) & \text{walking},
    \end{cases}
    \end{aligned}
\label{eq:motion_generation}
\end{equation}
where $\vect{p}_t$ and $\vect{\theta}_t$ are the torso position and orientation (unit quaternion), $\vect{v}_t$ and $\vect{\omega}_t$ are the linear and angular torso velocities, $\vect{q}_t$ and $\dot{\vect{q}}_t$ are the joint positions and velocities, and $c^L_t$ and $c^R_t$ are the left and right foot contact indicators. Hats $\hat{\cdot}$ denote target quantities. For walking, $f(\cdot)$ additionally includes the gait phase variable $\phi_t$. 
To ensure robustness and broad applicability, the control input $\vect{g}_t$ is randomized across its full range during training.

\subsection{Policy}
\label{sec:policy}
Actions $\vect{a}_t$ are position targets for Proportional-Derivative (PD) controllers at the joints. The robot's proprioceptive state is
\begin{equation}
    \vect{s}_t \coloneq (\vect{p}^\mathcal{P}_t, {\vect{\theta}}^\mathcal{P}_t, {\vect{v}}^\mathcal{R}_t, {\vect{\omega}}^\mathcal{R}_t,  \vect{q}_t, \dot{\vect{q}}_t, \vect{a}_{t-1}, \vect{a}_{t-2}, \vect{T}_t, \phi_t ),
\end{equation}
where $\vect{p}^\mathcal{P}_t$ and ${\vect{\theta}}^\mathcal{P}_t$ form the root pose relative to the path frame, and ${\vect{v}}^\mathcal{R}_t$ and ${\vect{\omega}}^\mathcal{R}_t$ denote the torso velocities expressed in the root frame. We also append the joint positions $\vect{q}_t$, joint velocities $\dot{\vect{q}}_t$, the actions of the two previous time steps,  $\vect{a}_{t-1}$ and $\vect{a}_{t-2}$, and the temperature $\vect{T}_t$ of the actuators. For walking, the policy is additionally conditioned on the gait phase variable $\phi_t$.

\subsection{Reward Formulation}
\label{sec:reward}

The reward includes four components: \emph{imitation}, \emph{regularization}, \emph{limits}, and \emph{impact reduction},
\begin{equation}
r_t = r^{\text{imitation}}_t + r^{\text{regularization}}_t + r^{\text{limits}}_t + r^{\text{impact reduction}}_t.
\label{eq:reward}
\end{equation}

The \emph{imitation} and \emph{regularization} terms follow standard practice~\cite{peng2018deepmimic,grandia2024bdx} and encourage accurate tracking of the reference motion with action penalties. The \emph{limit} terms capture constraints arising from Olaf’s compact mechanical design. The \emph{impact reduction} term reduces foot impacts and thereby significantly lowers footstep noise.

A detailed breakdown of the reward terms is provided in~\tabref{tab:rewards} with reward weights reported in~\tabref{tab:rewards_weights}, where hats $\hat{\cdot}$ denote target quantities derived from $\vect{x}_t$. The time index $t$ is omitted for readability. Below, we describe the reward functions in more detail.


\begin{table}[tb]
\begin{center}
    \caption{\textbf{Reward Terms.} Hats \(\hat{\cdot}\) denote reference quantities. Control-barrier rewards enforce thermal and joint limits, while penalizing large foot velocity rates reduces impact noise. $\boxminus$ is the $SO(3)$ log-map orientation difference and $\mathds{1}[\cdot]$ is the indicator function.}
    \footnotesize
    \begin{tabular}{l | l}\toprule
    \textbf{Name}                  &  \textbf{Reward Term} \\
    \midrule
    \multicolumn{2}{c}{\textit{Imitation}}  \\
    \midrule
    Torso position xy     &  $\exp\left(-200.0 \cdot \lVert \vect{p}_{x,y} - \hat{\vect{p}}_{x,y} \rVert_2^2 \right)$ \\
    Torso orientation    &  $\exp\left(-20.0 \cdot \lVert \vect{\theta} \boxminus \hat{\vect{\theta}} \rVert _2 ^2\right)$ \\
    Linear vel. xy  &  $\exp\left(-8.0 \cdot \lVert \vect{v}_{x,y} - \hat{\vect{v}}_{x,y} \rVert _2 ^2\right)$ \\
    Linear vel. z  &  $\exp\left(-8.0 \cdot \lVert \vect{v}_z - \hat{\vect{v}}_z \rVert _2 ^2\right)$ \\
    Angular vel. xy  &  $\exp\left(-2.0 \cdot \lVert \vect{\omega}_{x,y} - \hat{\vect{\omega}}_{x,y} \rVert _2 ^2 \right)$ \\
    Angular vel. z  &  $\exp\left(-2.0 \cdot \lVert \vect{\omega}_z - \hat{\vect{\omega}}_z \rVert _2 ^2 \right)$ \\
    Leg joint pos.     &  $-\lVert \vect{q}_l - \hat{\vect{q}}_l \rVert_2 ^2 $ \\
    Neck joint pos.    &  $-\lVert \vect{q}_n - \hat{\vect{q}}_n \rVert_2 ^2 $ \\
    Leg joint vel.     &  $-\lVert \dot{\vect{q}}_l - \hat{\dot{\vect{q}}}_l \rVert_2 ^2 $ \\
    Neck joint vel.    &  $-\lVert \dot{\vect{q}}_n - \hat{\dot{\vect{q}}}_n \rVert_2 ^2 $ \\
    Contact     &   $ \sum_{i\in\{L,R\}} \mathds{1}[c_i =\, \hat{c}_i]$ \\
    Survival             &  $1.0$ \\
    \midrule
    \multicolumn{2}{c}{\textit{Regularization}}  \\
    \midrule
    Joint torques     &  $-\lVert \vect{\tau} \rVert _2 ^2 $ \\
    Joint acc.       &  $-\lVert \ddot{\vect{q}} \rVert _2 ^2 $ \\
    Leg action rate  &  $-\lVert \vect{a}_{l} - \vect{a}_{t-1,l} \rVert _2 ^2 $ \\
    Neck action rate &  $-\lVert \vect{a}_{n} - \vect{a}_{t-1,n} \rVert _2 ^2 $ \\
    Leg action acc.  &  $-\lVert \vect{a}_{l} - 2\vect{a}_{t-1,l} + \vect{a}_{t-2,l} \rVert _2 ^2 $ \\
    Neck action acc. &  $-\lVert \vect{a}_{n} - 2\vect{a}_{t-1,n} + \vect{a}_{t-2,n} \rVert _2 ^2 $ \\
    \midrule
    \multicolumn{2}{c}{\textit{Limits}}  \\
    \midrule
    Neck temperature & $-\lVert \min (-\dot{\vect{T}}_n + \gamma_T (\vect{T}_{\max} -\vect{T}_n),\vect{0})\rVert_{1}$\\
    Joint limits (lower) & $-\lVert \min\left( \dot{\vect{q}} + \gamma_q(\vect{q}-(\vect{q}_{\min} + q_{m})),\vect{0} \right) \rVert_{1}$ \\
    Joint limits (upper) & $-\lVert \min\left( -\dot{\vect{q}} + \gamma_q((\vect{q}_{\max} - q_{m})-\vect{q}), \vect{0} \right) \rVert_{1}$ \\
    Foot–Foot collision &  $-\mathds{1}[\text{contact}(L,R)]$ \\
    \midrule
    \multicolumn{2}{c}{\textit{Impact Reduction}}  \\
    \midrule
    Sound suppression  & $-( \sum_{i \in \{\mathrm{L},\mathrm{R}\}} \min ({\Delta v}_{i,z}^{2},\; {\Delta v}_{\max}^2))$ \\
    \bottomrule
    \end{tabular}
\label{tab:rewards}
\end{center}
\end{table}

The reward $r^{\text{imitation}}_t$ incentivizes accurate imitation of the kinematic reference motion. Joint weights differ between the neck and legs to account for their substantially different reflected inertia. 
We apply early termination when the head, torso, upper legs, or arms are in contact with the ground. We additionally add regularization terms \(r_t^{\text{regularization}}\) to penalize excessive joint torques and encourage smooth actions, thereby reducing vibrations and unnecessary effort.

The reward $r^{\text{limits}}_t$ penalizes violations of critical physical constraints. We incorporate a temperature reward term as detailed in~\secref{sec:temp_model}, and implement joint limit penalties, described in~\secref{sec:joint_model}. Finally, to prevent self-collisions between Olaf’s two snowball feet, we include a penalty for foot–foot contact.

The term $r^{\text{impact reduction}}_t$ penalizes changes in velocity along the gravity direction between simulation steps, thereby producing smoother foot motions and reducing impact noise. Because the physics engine can generate large velocity changes during contact resolution, which may result in a large reward term and destabilize critic learning, we saturate the impact-reduction reward term in the reward computation.

\subsection{Thermal Modeling}
\label{sec:temp_model}
Olaf’s slim, costume-covered neck requires small actuators to support a heavy head, which caused frequent overheating in early experiments. We address this by requiring the actuator temperature $T$ to stay below a maximum temperature $T_{\max}$, as formalized through the inequality constraint in~\eqref{eq:temp_barrier}. We transform this constraint, which depends on the slowly varying temperature state, into a Control Barrier Function (CBF, \cite{ames2019control}) condition~\eqref{eq:temp_cbf}. The CBF condition shapes the desired behavior locally in time by imposing constraints on the time derivative.
Intuitively, it ensures that as the temperature approaches or exceeds the maximum, the system responds by maintaining $\dot{T} \le 0$, thereby preventing overheating
\begin{subequations}
\label{eq:temp_reward}
\begin{align}
    h_T(T) = T_{\max} - T &\ge 0, \label{eq:temp_barrier} \\
    \dot{h}_T(T) + \gamma_T h_T(T) &\ge 0, \quad \gamma_T > 0, \label{eq:h_cbf} \\
    -\dot{T} + \gamma_T (T_{\max} - T) &\ge 0, \quad \gamma_T > 0. \label{eq:temp_cbf}
\end{align}
\end{subequations}
The set of CBF constraints per actuator is translated into a penalty by computing the total violation, as defined in~\tabref{tab:rewards}. To implement the thermal CBF in simulation, we require a model of the actuator thermal dynamics. These dynamics are dominated by electrical Joule heating $P$, which scales with $\tau^2$ since torque $\tau \propto I$ and $P \propto I^2$. We therefore model the temperature as a first-order system driven by squared torque
\begin{equation}
\dot{T} = -\alpha (T - T_{\text{ambient}}) + \beta \tau^2,
\label{eq:temp_model}
\end{equation}
neglecting mechanical heat generation. The parameters $\alpha$, $\beta$, and $T_{\text{ambient}}$ are fitted from data, as detailed in~\secref{sec:implementation_details_thermal}. 

\subsection{Joint Limits}
\label{sec:joint_model}

To prevent joint-limit violations, we use a similar reward function based on CBF conditions~\eqref{eq:joint_cbf_condition}, which enforces a margin $q_{\mathrm{m}}$ from each joint's physical limits $q_{\min}$ and $q_{\max}$. For each joint, we define
\begin{equation}
h_q(q) = 
    \begin{cases}
    q - (q_{\min} + q_{\mathrm{m}}) & \ge 0, \quad \text{lower limit},\\
    (q_{\max} - q_{\mathrm{m}}) - q & \ge 0, \quad \text{upper limit},
    \end{cases}
\end{equation}
with the corresponding per-joint CBF constraints
\begin{equation}
    \begin{aligned}
    \dot{q} + \gamma_q \left(q - (q_{\min} + q_{\mathrm{m}})\right) \ge 0, \quad \text{lower limit},\\
    -\dot{q} + \gamma_q \left((q_{\max} - q_{\mathrm{m}}) - q \right) \ge 0, \quad \text{upper limit},
    \end{aligned}
    \label{eq:joint_cbf_condition}
\end{equation}
where $\gamma_q > 0$. We choose a margin of $q_{\mathrm{m}} = \SI{0.1}{\radian}$ and set $\gamma_q = 20$.

\section{Show Functions}
\label{sec:show_functions}

Olaf's show functions --- the eyes and eyebrows, the jaw mechanism, and the arms --- have low inertia and therefore minimally affect system dynamics.  
For this reason, they are separated from the main articulated backbone and controlled using classical methods.
To control Olaf's show functions, we must map their functional space, which is the space in which motions are animated and composited, to actuator space. This mapping is derived using a forward-kinematics solver~\cite{schumacher2021versatile} by uniformly sampling the functional region of interest and fitting a polynomial.

For Olaf's eyes, the functional space includes left and right eye yaw, coupled eye pitch, and eyelid closure. For the given eye mechanism, a first-order polynomial per actuator is sufficiently accurate. Olaf's left and right arms are implemented as spherical 5-bar linkages, each driven by two actuators. We parameterize their functional coordinates using two serial revolute angles: \emph{arm swing} followed by \emph{arm pitch}. Arm swing maps directly to the first actuator, whereas arm pitch is coupled through both actuators. The second actuator position is obtained through a cubic polynomial fit. After applying the actuator mapping, all eye and arm actuators are controlled using a PD loop.

For the jaw, the costume introduces significant external forces through fabric tension when the mouth is closed and wrinkling when it opens, which degrade tracking performance. To compensate for these effects, we add a feedforward term $\tau^{\text{jaw}}_{\text{ff}}(q^{\text{jaw}})$ to the PD controller.
We estimate this feedforward term by measuring the torque required to hold a set of uniformly sampled jaw angles $q^{\text{jaw}}$ across the full range of motion. Using least squares, we fit a first-order polynomial with an additional cosine term that captures the non-linearity observed in the data
\begin{equation}
    \tau^{\text{jaw}}_{\text{ff}}(q^{\text{jaw}})
    =  c_0 + c_1 q^{\text{jaw}} + c_{\text{cos}} \cos(q^{\text{jaw}}),
\end{equation}
where $c_0$, $c_1$, and $c_{\text{cos}}$ are the fitted model parameters.

\section{Implementation Details}

\label{sec:implementation_details}

\subsection{Thermal Model and Reward}
\label{sec:implementation_details_thermal}
We fit the parameters of the thermal model $(\alpha, \beta, T_{\mathrm{ambient}})$ in~\eqref{eq:temp_model} using a least-squares regression applied to an explicit Euler discretization of the thermal dynamics, sampled at \SI{50}{\hertz} to match the policy rate. The regression is performed on \SI{20}{\minute} of recorded data. A quantitative evaluation of the resulting model is provided in~\secref{sec:thermal_modeling}.
All parameters used in the thermal reward model are summarized in~\tabref{tab:thermal_model}.
The maximum actuator temperature, $T_{\text{max}}$, is set to \SI{80}{\degreeCelsius}, a value determined experimentally to prevent overheating while maintaining a conservative safety margin.

To limit the temperature range the policy needs to be exposed to during training, we clip the temperatures used in policy observations and rewards to the interval $[T_{\text{clip,min}}, T_{\text{clip,max}}]$. 
To ensure that the constraint in~\eqref{eq:temp_cbf} remains feasible at the upper clipping boundary, we choose $\gamma_T$ such that the constraint is satisfied under the smallest possible heat generation, i.e., when $\tau^2 = 0$. 

\begin{table}[tb]
\begin{center}
    \caption{\textbf{Thermal Model and Reward Parameters.} Neck actuator thermal model coefficients and reward parameters, specifying temperature dynamics, allowed limits, and the control-barrier coefficient.}
    \footnotesize
    \begin{tabular}{l c | l c}
    \toprule
    \multicolumn{2}{c|}{\textbf{Thermal Model}} & \multicolumn{2}{c}{\textbf{Reward Function}} \\
    \midrule
    $\alpha$ & $0.038$ & $T_{\max}$ & \SI{80}{\degreeCelsius} \\
    $\beta$ & $0.377$ & $T_{\text{clip}}$ & [\SI{70}{\degreeCelsius}, \SI{85}{\degreeCelsius}] \\
    $T_{\text{ambient}}$ & $43.94$ & $\gamma_T$ & $0.312$ \\
    \bottomrule
    \end{tabular}
\label{tab:thermal_model}
\end{center}
\end{table}

\subsection{Training}

We train the policies using Proximal Policy Optimization (PPO)~\cite{schulman2017proximal}. 
The critic receives privileged information, including noiseless measurements, friction parameters, and terrain height samples. We add noise to the actor observations and apply randomized disturbance forces during training. Both actor and critic are implemented as three-layer MLPs with $512$ units per layer. 
All reward weights are listed in \tabref{tab:rewards_weights}.

Training is performed in Isaac Sim, running $8192$ environments in parallel on a single RTX~4090 GPU. Policies are trained for $100$k iterations (approximately \SI{2}{\day}). 

\begin{table}[tb]
\centering
\caption{\textbf{Reward Weights.} Weights used for the standing and walking policies. Two values indicate \emph{Standing / Walking}; a single value applies to both.}
\footnotesize
\begin{tabular}{l | c | l | c}
\toprule
\shortstack{\textbf{Reward}\\\textbf{Name}} & \shortstack{\textbf{Standing}\\\textbf{/Walking}} &
\shortstack{\textbf{Reward}\\\textbf{Name}} & \shortstack{\textbf{Standing}\\\textbf{/Walking}} \\
\midrule \midrule

Torso position xy & $1.0 / 4.0$ & Neck action rate     & $5.0 / 10.0$ \\
Torso orientation  & $2.0 / 1.5$ & Leg action rate      & $2.0 / 5.0$  \\
Linear vel. xy & $1.5 / 2.5$ & Leg action acc.      & $0.5 / 1.0$  \\
Linear vel. z  & $1.0$ & Neck action acc.     & $15.0 / 10.0$ \\
Angular vel. z & $1.5$ & Neck temperature     & $2.0$ \\
Leg joint pos. & $15.0$ & Joint limits       & $0.5 / 0.2$ \\
Neck joint pos.& $40.0$ & Foot–Foot collision & $10.0 $ \\
Leg joint vel. & $1.0\cdot10^{-3}$ & Impact reduction & $2.5\cdot10^{-3}$ \\
Neck joint vel.& $0.5$  & Joint torques       & $1.0\cdot10^{-3}$ \\
Contact        & $2.0 / 1.0$  & Joint acc.          & $2.5\cdot10^{-6}$ \\
Survival       & $20.0$ & & \\
\bottomrule
\end{tabular}
\label{tab:rewards_weights}
\end{table}

\subsection{Runtime}
\label{sec:runtime}

After training, the policy networks are frozen and deployed to the robot's on-board computer. The proprioceptive state $\vect{s}_t$ is estimated using a state estimator that fuses IMU and actuator measurements~\cite{hartley2020contact}. The policy runs at \SI{50}{\hertz}, and its output is upsampled to the \SI{600}{\hertz} actuator rate using a first-order hold (i.e., linear interpolation between successive actions). Finally, the upsampled actions are passed through a low-pass filter with a cut-off of \SI{37.5}{\hertz} to ensure smooth motor commands.

Using the architecture from \cite{grandia2024bdx}, puppeteering commands are processed by an \emph{Animation Engine}, which blends triggered content and maps real-time puppeteering inputs to policy control input $\vect{g}_t$, show function signals, and audio triggers through a three-stage process:

\begin{itemize}
  \item \textbf{Background animations}: plays looped whole-body animations to introduce subtle idle behaviors such as eye saccades and arm adjustments.
  \item \textbf{Triggered animations}: layers short animated clips (e.g., gestures or spoken lines) on top of the background animations.
  \item \textbf{Joystick-driven control}: adjusts the output pose based on teleoperation inputs. During standing, it controls gaze and body posture; during walking, it controls gaze and path velocity.
\end{itemize}

\section{Results}

We evaluate Olaf's mechatronic design in the real world, with its performance demonstrated directly in the visual results. The asymmetric leg design enables faithful imitation of Olaf's characteristic animations while adhering to all design requirements. The magnetic arms and nose enable in-character gags, as highlighted in the supporting video.

In the following, we analyze the robot’s control mechanisms in detail.

\subsection{Tracking Performance}
\label{sec:tracking}

We compare the kinematic reference to the executed motion on the robot (see supplementary video). To quantify this, we evaluate the mean absolute joint-tracking error over the full trajectory and all joints of our walking and standing policies across the full range of control inputs $\vect{g}_t$ for \SI{5}{\minute} each. For the standing policy, we achieve an error of \SI{3.87}{\degree}~$\pm$~\SI{2.40}{\degree}, and for the walking policy, \SI{4.02}{\degree}~$\pm$~\SI{2.01}{\degree}.

We additionally analyze the effect of Olaf's characteristic heel–toe walking by training a policy without it to assess its effect on the visual appearance of his gait. As shown in the video, the resulting motion appears more robotic.

\subsection{Thermal Model Evaluation}
\label{sec:thermal_modeling}

\begin{figure}[t]
    \centering
    \includegraphics[width=\linewidth]{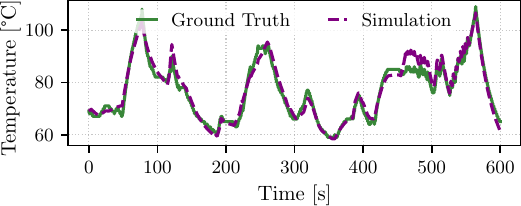}
    \caption{\textbf{Thermal Model.} Validation of the thermal model in~\eqref{eq:temp_model}, comparing \textcolor{magenta}{predicted} actuator temperatures, $T$, with \textcolor{green}{measured} values over a \SI{10}{\minute} rollout. Temperatures are quantized as reported by the actuator.}
    \label{fig:temp_model}
\end{figure}

As shown in~\figref{fig:temp_model}, we evaluate the thermal model fitted with the parameters in~\tabref{tab:thermal_model} by simulating the temperature evolution over an unseen \SI{10}{\minute} trajectory, initialized to the measured value at time~$0$. Across the full trajectory, the model achieves a mean absolute error of \SI{1.87}{\degreeCelsius}, demonstrating good predictive accuracy.

The timescale of the thermal model is far slower than that of the balancing dynamics, making it a challenging RL objective. To evaluate the effectiveness of the proposed thermal reward, we compare the temperature evolution of the neck-pitch actuator, identified in our experiments as the actuator most prone to overheating, and the mean absolute joint-tracking error for policies trained with and without the thermal reward. The evaluation is performed on a trajectory where Olaf looks upward and follows a predefined motion, slowly turning his head left and right. 

\begin{figure}[t]
    \centering
    \includegraphics[width=\linewidth]{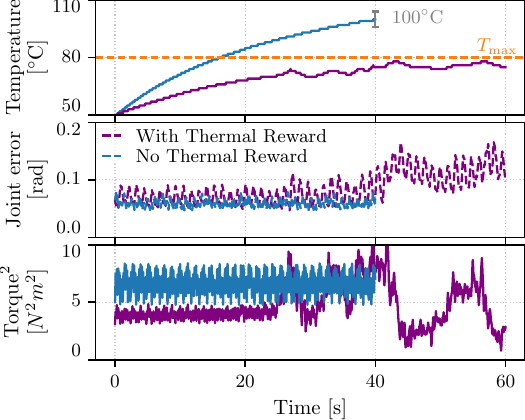}
    \caption{\textbf{Thermal Reward Evaluation.} Neck-pitch temperature, mean absolute joint-tracking error, and squared neck-pitch torque for policies trained \textcolor{magenta}{with} and \textcolor{blue}{without} the thermal reward. The thermal reward slows temperature rise while maintaining tracking at low temperatures and slightly relaxes tracking near the temperature limit to prevent overheating.}
    \label{fig:temp_reward}
\end{figure}

As shown in \figref{fig:temp_reward}, the baseline policy without the thermal reward causes the actuator temperature to rise rapidly, reaching \SI{100}{\degreeCelsius} within \SI{40}{\second}, at which point the experiment was stopped to prevent actuator damage. The policy with the thermal reward exhibits only a slightly larger tracking error, while the actuator temperature rises significantly slower. This is explained by the squared torque, which shows that the policy reduces torque usage well before reaching the temperature limit, while maintaining nearly the same tracking accuracy. As the temperature approaches the \SI{80}{\degreeCelsius} threshold, the squared torque increases as the policy adjusts the head toward a more horizontal orientation. This requires more torque initially, but lowers torque demand over time. To evaluate long-duration performance, we executed the same motion continuously for \SI{1}{\hour}, during which the system reached a thermal equilibrium. Over the final minute of the experiment, we measured a mean actuator temperature of \SI{77.3}{\degreeCelsius} and a mean absolute joint tracking error of \SI{0.14}{\radian}. The video further illustrates this behavior, showing how the policy gradually relaxes tracking near the temperature limit.

\subsection{Foot Impact}
\label{sec:eval_foot_impact_reduction}
We evaluate the foot impact reduction reward for sound suppression by comparing its addition to the baseline on hardware. Over a \SI{5}{\minute} run, this reward reduces the mean sound level by \SI{13.5}{\decibel}, as clearly noticeable in the video. Despite the reward, tracking performance and similarity to the kinematic reference are largely preserved.

\begin{figure}[t]
    \centering
    \includegraphics[width=\linewidth]{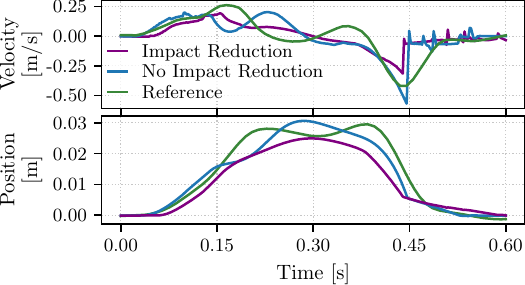}
    \caption{\textbf{Foot Impact Reduction.} Comparison of z-foot velocity and position for the \textcolor{green}{reference} and policies trained \textcolor{magenta}{with} and \textcolor{blue}{without} the foot impact reduction reward.}
    \label{fig:foot_impact_vel}
\end{figure}

\figref{fig:foot_impact_vel} shows the vertical foot velocity and position profiles for the reference, as well as policies trained with and without the foot impact reduction reward during a swing phase. The reward acts as a regularizer: the overall trajectory is preserved, but small nuances of the reference, such as the mid-swing dip of the foot, are smoothed out. The no-impact-reduction policy imposes less smoothing, but exhibits higher peak velocities at foot impact. This demonstrates that the reward effectively reduces impact forces while maintaining the overall motion profile.

\subsection{System Deployment}

We have validated the robot through extended deployments and real-world testing. The thermal reward prevented actuator overheating, with no noticeable detrimental effect on the performances.

\section{Conclusion}

This work presented Olaf, a freely walking robot that accurately imitates the animated character, which is challenging to physically represent, in terms of style and appearance. We hope this work inspires others to push the boundaries beyond standard bipedal and quadrupedal robotics.

We addressed challenging design requirements by proposing an asymmetric 6-DoF leg mechanism hidden beneath a foam skirt and the integration of remotely-actuated linkages. We tackled control requirements by using RL and impact-reducing rewards to significantly reduce stepping sound. We further incorporated control barrier function constraints to mitigate actuator overheating with a thermal model and to prevent joint-limit violations. 

Although the results are specific to Olaf, we expect our impact-reducing rewards to decrease wear and extend a robot's operational lifetime. In our experiments, the CBF reward formulation has prevented constraint violations, providing an effective tool for handling constraints within an RL context. CBF rewards are particularly useful for long-horizon effects such as thermal management, enabling a robot to self-manage temperature via learned policies.

Several open research directions remain. A higher fidelity thermal model could capture contributions of mechanical effects like friction, or gradual heating of actuator enclosures during extended operation. In this work, interaction forces between the costume and legs were handled through domain-randomized disturbance forces. Explicitly modeling these effects could reduce reliance on randomization and provide more targeted training. 

\addtolength{\textheight}{-3.5cm}   




\bibliography{bibliography.bib}
\bibliographystyle{IEEEtran}

\appendices

\end{document}